%% file: neurips_camera_ready.tex
\documentclass{article}

\PassOptionsToPackage{square,numbers}{natbib}

\usepackage[final]{neurips_2023}

\usepackage{algorithm}
\usepackage{algpseudocode}
\usepackage[utf8]{inputenc} 
\usepackage[T1]{fontenc}    
\usepackage{hyperref}       
\usepackage{url}            
\usepackage{booktabs}       
\usepackage{amsfonts}       
\usepackage{nicefrac}       
\usepackage{microtype}      
\usepackage{xcolor}         
\usepackage{mathtools}
\usepackage{enumitem}

\usepackage{caption}
\usepackage{subcaption}
\usepackage{amsthm}
\usepackage{verbatim}
\usepackage{cases}
\newtheorem{prop}{Proposition}

\newtheorem{Lemma}{Lemma}

\newtheorem{defi}{Definition}
\theoremstyle{remark}

\newcommand{\R}{\mathbb{R}}
\newcommand{\E}{\mathbb{E}}

\newcommand{\N}{\mathbb{N}}

\newcommand{\eps}{\varepsilon} 

\title{Jarzynski Reweighting and Sampling Dynamics for Training Energy-Based Models: Theoretical Analysis of Different Transition Kernels}

\author{%
  Davide Carbone \\
  Dipartimento di Scienze Matematiche,
  Politecnico di Torino, Italy\\
  Istituto Nazionale di Fisica Nucleare, Sezione di Torino, Italy\\
  Laboratoire de Physique de l’École normale supérieure, ENS, Université PSL, France\\
  \texttt{davide.carbone@polito.it} \\
}

\begin{document}

\maketitle

\begin{abstract}
 Energy-Based Models (EBMs) provide a flexible framework for generative modeling, but their training remains theoretically challenging due to the need to approximate normalization constants and efficiently sample from complex, multi-modal distributions. Traditional methods, such as contrastive divergence and score matching, introduce biases that can hinder accurate learning. In this work, we present a theoretical analysis of Jarzynski reweighting, a technique from non-equilibrium statistical mechanics, and its implications for training EBMs. We focus on the role of the choice of the kernel and we illustrate these theoretical considerations in two key generative frameworks: (i) flow-based diffusion models, where we reinterpret Jarzynski reweighting in the context of stochastic interpolants to mitigate discretization errors and improve sample quality, and (ii) Restricted Boltzmann Machines, where we analyze its role in correcting the biases of contrastive divergence. Our results provide insights into the interplay between kernel choice and model performance, highlighting the potential of Jarzynski reweighting as a principled tool for generative learning.
\end{abstract}

\section{Introduction}
\label{sec:intro}
Probabilistic models are at the core of modern generative artificial intelligence and unsupervised learning, providing a framework to both describe observed data and generate new samples. A wide range of approaches have been developed, including variational autoencoders, generative adversarial networks, normalizing flows, and diffusion-based models. Among these, energy-based models (EBMs)~\cite{lecun2006tutorial,gutmann2010noise,song2020sliced} stand out due to their deep connections with statistical physics. EBMs define probability distributions through a energy function $U$, leading to the unnormalized density  $\rho = \exp(-U) / Z$ , where $Z$ is a normalization constant. This formulation allows for flexible modeling but presents a major challenge: training requires approximating expectations over  $\rho$, which depends on sampling from a typically high-dimensional and multimodal distribution -- generally an hard task. However, once properly trained, an EBM provides a normalized probability distribution over data points, making it inherently {\it interpretable}, as it assigns meaningful likelihoods to different regions of the data space.\\
Because of its relation with sampling, a key difficulty in training EBMs lies in properly capturing multimodal data distributions where different regions of the energy landscape are weighted unequally. In real-world datasets, this often translates to modes of varying importance, and an incorrect estimation of their relative weights can lead to biased learning outcomes. This issue is particularly severe in applications where fairness considerations are crucial \cite{mehrabi2021survey}.

Two principal approaches are commonly used to train EBMs. One relies on score matching, which minimizes the Fisher divergence by learning the gradient of the energy function,  $\nabla U_\theta(x)$. It is a sampling-free approach and computationally efficient due to techniques such as denoising score matching~\cite{hyvarinen2005estimation}, this approach inherently struggles with mode imbalance, as it does not properly estimate the normalization factor $Z$ and, consequently, the relative probabilities of different regions in the state space. The second approach minimizes the cross-entropy between the data and model distributions, which does account for mode weights but requires generating unbiased samples from the model. This typically involves running Markov Chain Monte Carlo methods, such as the unadjusted Langevin algorithm, until convergence—an often impractical requirement in high-dimensional, non-log-concave settings. As a result, approximations such as contrastive divergence (CD) or persistent contrastive divergence (PCD) \cite{hinton2002training,welling2002new,carreira2005contrastive} are widely used, but these introduce biases that degrade the quality of learning, particularly in the presence of sharp energy barriers or poor mixing \cite{hyvarinen2007connections}.\\
To address these challenges the concept of Jarzynski reweighting has been recently proposed for training Energy-Based Models. Jarzynski’s equality allows the computation of free energy differences between distributions using out-of-equilibrium dynamics, without requiring equilibrium sampling. This method has been extended to EBMs in works like \cite{carbone2024efficient, carbone2024generative}, where sequential Monte Carlo methods are used to directly train EBMs by approximating the gradient of the cross-entropy objective more efficiently. By incorporating {\it Jarzynski reweighting}, this approach overcomes issues with traditional methods like MCMC and contrastive divergence, which can be computationally expensive and prone to bias, especially in high-dimensional, multi-modal distributions.

\paragraph{Main contributions.} In this work, we extend the use of Jarzynski reweighting to two important classes of generative models, demonstrating its versatility in different sampling frameworks:

\begin{itemize}[leftmargin=0.2in]
\item Flow-based diffusion models: We show that Jarzynski reweighting can be used to correct numerical errors introduced by discretization during the generative phase. In this context, we reinterpret the method in relation to stochastic interpolants, providing a principled approach to improving sample quality.
\item Restricted Boltzmann Machines (RBMs): We apply Jarzynski reweighting to the training of RBMs, offering a correction to the biases introduced by traditional contrastive divergence-based methods. This should allowa for a more accurate estimation of the cross-entropy gradient, leading to improved learning dynamics.
\end{itemize}

\paragraph{Related works.} Training Energy-Based Models remains a longstanding challenge in machine learning, with comprehensive overviews available, such as those in \cite{lecun2006tutorial, song2021train}. The introduction of modern deep neural network architectures for modeling energy functions was proposed in \cite{xie2016theory}, and a connection between EBMs and classifiers is discussed in \cite{grathwohl2019your}. Sampling from unnormalized probability distributions, another central problem in statistical modeling, has been well explored, with foundational works found in \cite{brooks2011handbook, liu2001monte}.

Score-matching methods and their variants, first explored in \cite{hyvarinen2005estimation, vincent2011connection, swersky2011autoencoders}, have faced challenges in EBM training, particularly in addressing the presence of multiple, imbalanced modes in the target distribution. These shortcomings are examined in \cite{wenliang2022failure,wenliang2019learning, song2019generative}. Contrastive divergence, introduced in \cite{hinton2002training, welling2002new, carreira2005contrastive}, is a widely used technique for approximating the gradient of the log-likelihood. However, CD algorithms are known for their biases due to incomplete sampling steps, which are difficult to eliminate. Persistent Contrastive Divergence, proposed in \cite{tieleman2008training}, overcomes this issue by removing restarts and evolving the walkers using the Unadjusted Langevin Algorithm. Nevertheless, these methods still suffer from uncontrolled gradient biases that are hard to mitigate. There have been several attempts to refine these methods, often in combination with other unsupervised learning techniques, as seen in \cite{xie2018cooperative, nijkamp2019learning, gao2020flow}.

The Jarzynski equality, introduced in \cite{jarzynski1997nonequilibrium}, provides a direct relationship between the normalizing constants (or free energy) of two distributions connected by out-of-equilibrium continuous-time dynamics. A discrete-time version of JE is utilized in Neal’s annealed importance sampling \cite{neal2001annealed}, a method belonging to the broader class of Sequential Monte Carlo techniques \cite{doucet2001sequential}. These methods have been applied in the context of generative models, including variational autoencoders \cite{anh2018autoencoding, ding2020learning}, normalizing flows \cite{midgley2023flow, du2023reduce}, and diffusion models \cite{pmlr-v37-sohl-dickstein15, doucet2022scorebased}. 

Regarding generative models, Restricted Boltzmann Machines are a type of EBM where the energy is defined over binary visible and hidden units, with applications in unsupervised learning and dimensionality reduction \cite{salakhutdinov2009deep, hinton2012practical}. These models are typically trained using methods like contrastive divergence \cite{hinton2002training}. Gaussian RBMs extend the treatment to continuous data and are widely used in machine learning \cite{cho2011improved}. Flow-based methods, as for instance Flow Matching or Stochastic Interpolation, provide a powerful framework for learning complex distributions by modeling invertible transformations between simple distributions and the target distribution \cite{lipmanflow, albergo2023stochastic}. These techniques offer an alternative to standard variational methods, providing a flexible and powerful approach to model complex, multi-modal distributions.
\section{Energy-Based Models}
\label{sec:ebm}

\paragraph{Setup, notations, and assumptions.}
We consider the problem of estimating an unknown probability density function (PDF) \(\rho_*(x)\), given \(n\) data points \(\{x_i^*\}_{i=1}^n \subset \mathbb{R}^d\), using an energy-based model (EBM). Specifically, we seek a parametric energy function \(U_\theta: \mathbb{R}^d \to [0, \infty)\), \(\theta \in \Theta\), such that the associated Boltzmann-Gibbs density  
\begin{equation}   
\label{eq:ebm:def}
\rho_{\theta}(x)=Z_{\theta}^{-1} e^{-U_{\theta}(x)}, \quad Z_{\theta}=\int_{\mathbb{R}^d} e^{-U_{\theta}(x)}dx
\end{equation}
approximates \(\rho_*(x)\). The partition function \(Z_{\theta}\) is intractable, but EBMs allow sampling via MCMC methods, whose design is integral to model training.  

We assume \(U_\theta\) satisfies  
\begin{equation}
\label{eq:assump:U}
    \begin{gathered}
    U_\theta \in C^2(\R^d), \quad \exists L > 0: \|\nabla\nabla U_\theta(x)\|\le L \ \forall x;\\
    \exists a>0, \mathcal{C} \text{ compact}: x\cdot \nabla U_\theta(x) \ge a |x|^2 \ \forall x \notin \mathcal{C}.
\end{gathered}
\end{equation}
These ensure \(Z_\theta < \infty\) and ergodicity of Langevin-based samplers~\cite{oksendal2003stochastic,mattingly2002stochastic,talay1900expansion}. Importantly, \eqref{eq:assump:U} does not imply \(U_\theta\) is convex, allowing multimodal densities \(\rho_\theta\). This is especially true in deep generative models where $U_\theta$ is a neural network with learnable parameters $\theta$. 
\paragraph{Cross-entropy minimization.}
To measure the quality of the EBM and train its parameters one can use the cross-entropy of the model density $\rho_\theta$ relative to the target density $\rho_*$
\begin{equation} 
\label{2}
H(\rho_\theta,\rho_*) = \log Z_\theta + \int_{\R^d} U_\theta(x)\rho_*(x)dx.
\end{equation}
Using \(\partial_\theta \log Z_{\theta} = - \int_{\mathbb{R}^d} \partial_\theta U_{\theta}(x)\rho_{\theta}(x) dx\), its gradient is  
\begin{equation}
\label{4:0}
\begin{aligned}
    \partial_\theta H(\rho_{\theta},\rho_*)&=\E_*[\partial_\theta U_{\theta}]-\E_{\theta} [\partial_\theta U_{\theta}].
\end{aligned}
\end{equation}
This quantity can be exploited to perform gradient based minimization of CE, which is related to KL divergence and log-likelihood (e.g. cfr. \cite{carbphd}). However, it is harder to optimize, as estimating \(Z_\theta\) and \(\E_\theta[\partial_\theta U_{\theta}]\) is challenging. The reason is evident: estimating $\E_\theta$ requires samples {\it from the very same distribution} $\rho_\theta$ we are learning. Traditional methods like Contrastive Divergence and Persistent Contrastive Divergence \cite{hinton2002training,carreira2005contrastive} approximate \(\E_\theta[\partial_\theta U_{\theta}]\) but lack convergence guarantees and can be unstable. For reader's convenience we report in Appendix \ref{app:cd:pcd} well known results on this fact.

\section{Preliminaries: Jarzynski reweighting}
\label{sec:jarz}

In this section we recall the result found in \cite{carbone2024efficient,carbone2024generative} on EBM training via nonequilibrium statistical physics tools.  \\
\begin{defi} Let $(S, \mathcal{B}(S))$ be a measurable space, where $S$ is a state space and $\mathcal{B}(S)$ is its Borel $\sigma$-algebra. A Markovian transition kernel (or Markov kernel) is a function $\pi : \mathcal{B}(S) \times S \to [0,1]$ such that:

\begin{itemize}
    \item For every fixed $x \in S$, the function $\pi( \cdot,x)$ is a probability measure on $(S, \mathcal{B}(S))$.
    \item For every fixed $A \in \mathcal{B}(S)$, the function $\pi( A,\cdot)$ is a measurable function on $S$.
\end{itemize}
\end{defi}
In a nutshell, $\pi(X,Y)$ is the probability of $Y$ given that the previous state was $X$. In the following, we will need $\pi$ to be time dependent, and we denote this with $\pi_t$. We stress that markovianity is a state-related property, not influenced by any time dependence of $\pi$.
\label{sec:discrete}
\begin{prop}[Carbone, Hua, Coste, and Vanden-Eijnden, 2023 \cite{carbone2024efficient}]
\label{prop:3}
Assume that the parameters $\theta$ are evolved by some time-discrete protocol $\{\theta_k\}_{k\in \N_0}$ and that~\eqref{eq:assump:U} hold.  Given any $h\in(0,L)$ and any time dependent Markovian transition kernel $\pi_k : \mathbb{R}\times \mathcal{B}(\mathbb{R^d})\times\mathbb{R^d} \to [0,1]$, let $X_k\in \R^d$ and $A_k\in\R$ be given by the iteration rule
\begin{equation}
\label{eq:X:A:k}
\left\{ 
\begin{aligned}
    X_{k+1}&\sim \pi_k(X_{k+1},X_{k})&&X_0\sim\rho_{\theta_0}\\
    A_{k+1}&=A_k+U_{\theta_k}(X_k) -U_{\theta_k}(X_{k+1}) +\log \frac{ \pi_k(X_k,X_{k-1})}{\pi_{k-1}(X_{k-1},X_{k})},&&A_0=0,
    \end{aligned}\right.  
\end{equation}
where $U_{\theta}(x)$ is the model energy. Then, for all $k\in \N_0$,
\begin{equation}
\label{eq:grad:Z:k}
    \E_{\theta_k} [\partial_{\theta} U_{\theta_k}]=\frac{\E[  \partial_{\theta} U_{\theta_k}(X_k)e^{A_k} ]}{\E [ e^{A_k}]}, \qquad Z_{\theta_k} = Z_{\theta_0} \E \left[e^{A_k}\right]
\end{equation}
where the expectation on the left-hand side is with respect to $\rho_{\theta_k}$ and the expectations on the right-hand side are over the law of the joint process $(X_k,A_k)$.
\end{prop}

We provide an adaptation of the proof contained in \cite{carbone2024efficient,carbone2024generative}.
\begin{proof}
 By the definition of $A_k$, we have
\begin{equation}
\label{eq:X:A:k:s2}
\begin{aligned}
    \exp(A_k) &= e^{-U_{\theta_k}(X_k)+U_{\theta_0}(X_0)} \prod_{q=1}^k \frac{ \pi_q(X_q,X_{q-1})}{\pi_{q-1}(X_{q-1},X_{q})}
\end{aligned}
\end{equation}
The joint probability density function of the path $(X_0,X_1,\ldots,X_k)$ at any $k\in \N$ is
\begin{equation}
\label{eq:varrho}
    \varrho(x_0,x_1,\ldots, x_k) = Z_{\theta_0}^{-1} e^{-U_{\theta_0}(x_0)} \prod_{q=1}^k \pi_{q-1}(x_{q-1},x_{q})
\end{equation}
we deduce from~\eqref{eq:X:A:k:s2} and \eqref{eq:varrho} that, given an $f:\R^d \to \R$,  we can express the expectation $\E[f(X_k) e^{A_k}]$ as the integral
\begin{equation}
    \label{eq:expect:int}
    \begin{aligned}
    &\E[f(X_k) e^{A_k}]\\
    &= \int_{\R^{dk}} f(x_k) e^{-U_{\theta_k}(x_k)+U_{\theta_0}(x_0)} \prod_{q=1}^k \frac{ \pi_q(x_q,x_{q-1})}{\pi_{q-1}(x_{q-1},x_{q})} \varrho(x_0,x_1,\ldots, x_k) dx_0\cdots dx_k\\
    &  = Z_{\theta_0}^{-1} \int_{\R^{dk}} f(x_k) e^{-U_{\theta_k}(x_k)} \prod_{q=1}^k  \pi_q(x_q,x_{q-1})dx_0\cdots dx_k
    \end{aligned}
\end{equation}
Since $\int_{\R^d} \pi_k(x,y) dy =1$ for all $k\in\N_0$ and all $x\in\R^d$, we can perform the integrals over $x_0$, then $x_1$, etc. in this expression to be left with
\begin{equation}
    \label{eq:expect:int:2}
    \E[f(X_k) e^{A_k}]= Z_{\theta_0}^{-1} \int_{\R^{d}} f(x_k) e^{-U_{\theta_k}(x_k)} dx_k
\end{equation}
Setting $f(x)=1$ in this expression gives
\begin{equation}
    \label{eq:f1}
    \E[e^{A_k}]= Z_{\theta_0}^{-1} \int_{\R^{d}} e^{-U_{\theta_k}(x_k)} dx_k = Z_{\theta_0}^{-1} Z_{\theta_k}
\end{equation}
which implies the second equation in~\eqref{eq:grad:Z:k}; setting $f(x_k) = \partial_{\theta}U_{\theta_k}(x_k)$ in~\eqref{eq:expect:int:2} gives
\begin{equation}
    \label{eq:f2}
    \E[\partial_{\theta}U_{\theta_k}(X_k)e^{A_k}]= Z_{\theta_0}^{-1} \int_{\R^{d}} \partial_{\theta}U_{\theta_k}(x_k) e^{-U_{\theta_k}(x_k)} dx_k = Z_{\theta_0}^{-1} Z_{\theta_k} \E_{\theta_k} [\partial_{\theta}U_{\theta_k}]
\end{equation}
which can be combined with~\eqref{eq:f1} to arrive at the first equation in~~\eqref{eq:grad:Z:k}.\hfill 
\end{proof}
For completeness we also give a continuous-time version of this proposition in Appendix~\ref{app:continuous}, together with a connection with Jarzynski equality \cite{jarzynski1997nonequilibrium}.  We stress that the inclusion of the so-called Jarzyski weights $\exp(A_k)$ in~\eqref{eq:grad:Z:k} is key, as $\E [ \partial_{\theta} U_{\theta_k}(X_k)] \not \not = \E_{\theta_k} [\partial_{\theta} U_{\theta_k}]$ in general. We also stress that ~\eqref{eq:grad:Z:k} holds \textit{exactly} for $h>0$, without time discretization error terms. In summary,  Proposition~\ref{prop:3} shows that we can compute exactly any expectation with respect to $\rho_{\theta_k}$ for any choice of the protocol; hence, the problematic expectation on r.h.s. of \eqref{4:0} can be computed exactly. Having a precise estimate of $\partial_\theta H$ makes cross-entropy minimization possible without approximations present in standard approaches. That is, one can couple \eqref{eq:X:A:k} with the gradient-descent protocol $\theta_{k+1}=\theta_{k}-\partial_\theta H(\rho_{\theta_k},\rho_*)$ and train an EBM; the main difference with respect to contrastive divergence and similar approaches is that Proposition \ref{prop:3} provides also for free an estimate of the normalization constant $Z_{\theta_k}$. Having a normalized trained EBM corresponds to have a full probabilistic model that describes the data -- one can compute the probability, and not just energy, of any generated, or real, data point. From a statistical point of view, it corresponds to {\it interpretability}, i.e. to the ability of comparing data points using the trained EBM, which is a not trivial feature of a trained generative model. 
\section{Main results}
In this section we will elaborate on the choice of the transition kernel, that is on the dynamics that is chosen for $X_k$. One expects that in practical implementations such choice can be important, and theoretical clarifications in that context can be important. Firstly, we will analyze in details the case of Unandjusted Langevin Algorithm, i.e. the one adopted in \cite{carbone2024efficient}, in relation with stochastic interpolation. Secondly, we will discuss the structure of \eqref{eq:X:A:k} is case of Poissonian and Gaussian-Poissonian Restricted Boltzmann Machines.
\subsection{Unadjusted Langevin Algorithm and Stochastic Interpolation}
\label{sec:ula}
\begin{defi}
    The dynamics of ULA is associated to the transition kernel $\pi_k : \mathbb{R}\times \mathcal{B}(\mathbb{R}^d)\times\mathbb{R^d} \to [0,1]$
    \begin{equation}
    \pi_k(x,y)=(4\pi h)^{-d/2}\exp\Big(-\frac{1}{4h}\left|y-x+h\nabla U_{\theta_k}(x)\right|^2\Big)
\end{equation}
for $h>0$. That is, to the first order Euler-Maruyama integration of the following SDE
\begin{equation}
    dX_t=-\alpha \nabla U_{\theta(t)}(X_t) dt+\sqrt{2\alpha}\,dW_t
\end{equation}
where $\alpha>0$, $U_{\theta(t)}(x)$ is the model energy and $W_t\in\R^d$ is a standard Wiener process.
\end{defi}
The use of ULA for sampling was originally introduced by Parisi et al. \cite{parisi1981correlation} in the case of a fixed Boltzmann-Gibbs target ensemble, that is $\theta_k=const$ in the present notation. After a sufficient amount of iterations $K$ one expects to produce samples $X_K\sim e^{-U}/Z$ with a certain degree of approximation, cfr. \cite{durmus2017nonasymptotic}. Here we recall its application to EBM training as presented in \cite{carbone2024efficient}, as a Lemma of Proposizion \ref{prop:3}:
\begin{Lemma}(cfr. Proposition 1 in \cite{carbone2024efficient})
    Assume that the parameters $\theta$ are evolved by some time-discrete protocol $\{\theta_k\}_{k\in \N_0}$ and that~\eqref{eq:assump:U} hold. Given any $h\in(0,L)$, let $X_k\in \R^d$ and $A_k\in\R$ be given by the iteration rule
\begin{equation}
\left\{ 
\begin{aligned}
    X_{k+1}&=X_k-h \nabla U_{\theta_k}(X_k) +\sqrt{2h}\, \xi_k,\quad\quad\quad\quad &&X_0\sim \rho_{\theta_0},\\
    A_{k+1}&=A_k-\alpha_{k+1}(X_{k+1},X_{k})+\alpha_{k}(X_{k},X_{k+1}),&&A_0=0,
    \end{aligned}\right.  
\end{equation}
where $U_{\theta}(x)$ is the model energy, $\{\xi_k\}_{k\in \N_0}$ are independent $N(0_d,I_d)$,  and we defined
\begin{equation}
\label{eq:alpha}
    \alpha_k(x,y)= U_{\theta_k}(x) + \tfrac12 (y-x)\cdot \nabla U_{\theta_k}(x)+ \tfrac14 h |\nabla U_{\theta_k}(x)|^2
\end{equation}
\end{Lemma}
Here we present a generalization of such known result in presence of drift. Let us consider the following SDE for $\eps>0$
\begin{equation}
\label{eq:SDE-int}
    dX_t=[b(t,X_t)+\eps s(t,X_t)]dt+\sqrt{2\eps}\,dW_t
\end{equation}
where
 \begin{equation}
\label{1}
    s(t,x)=-\nabla U_{\theta(t)}(x)\coloneqq -\nabla U(t,x)
\end{equation}
and $X_t\sim \rho(x,t)=\exp(-U(t,x))/Z_t$. The vector fields $b(t,x)$ and $s(t,x)$ are respectively the drift and the score as defined in contexts as stochastic interpolation \cite{albergo2023stochastic} or, generally, flow-based generative models \cite{lipmanflow}, and the drift is such that $\nabla\cdot(b\rho)=\partial_t \rho$. Notice how the role of $b(t,x)$ is, in some sense, to substitute the Jarzynski correction --- as Jarzynski himself suggested \cite{vaikuntanathan2008escorted} before the advent of generative models, the estimation of free energy differences would be perfect if one knew the exact $b$ that {\it escorts} the population. The meaning of such velocity field is to keep the ensemble in equilibrium with Boltzmann-Gibbs density associated to the time dependent potential. In fact, we recall as the Fokker-Planck PDE associated to \eqref{eq:SDE-int} is
\begin{equation}
  \partial_t \rho +\nabla\cdot ((b+\varepsilon s)\rho-\varepsilon \nabla\rho) = 0
\end{equation} 
meaning that $\rho(x,t)=\exp(-U(t,x))/Z_t$ is a solution for the initial datum $\rho(x,0)=\exp(-U(0,x))/Z_t$. In continuous time there is no need for Jarzynski correction in presence of $b$.\\
The Euler-Maruyama time discretization of \eqref{1} yields
\begin{equation}
\label{3}
    X_{k+1}=X_k+[b(t_k,X_t)+\eps s(t_k,X_t)]h +\sqrt{2h\eps}\, \xi_k,\quad\quad\quad\quad X_0\sim \rho_{0},
\end{equation}
The integration of the exact SDE \eqref{3} in principle produce samples from $\rho(t,x)$ at each time, but numerical error and arbitrary noise scale $\eps$ can represent an issue in this sense. We show how using Jarzynski weights one can better control these two hyperparameters.
Before proceeding, we recall the shorthand $f_k(x)\coloneqq f(t_k,x)$ w.r.t. dependence in time.
\begin{prop}
\label{prop:4}
Assume that the parameters $\theta$ are evolved by some time-discrete protocol $\{\theta_k\}_{k\in \N_0}$ and that~\eqref{eq:assump:U} hold. Given any $h\in(0,L)$, let $X_k\in \R^d$ and $A_k\in\R$ be given by the iteration rule
\begin{equation}
\label{eq:X:A:k:bis}
\left\{ 
\begin{aligned}
    X_{k+1}&=X_k+[b(t_k,X_t)+\eps s(t_k,X_t)]h +\sqrt{2h\eps}\, \xi_k,\quad\quad\quad\quad &&X_0\sim \rho_{0},\\
    A_{k+1}&=A_k+\zeta_{k}(X_{k},X_{k+1})&&A_0=0,
    \end{aligned}\right.  
\end{equation}
where $\{\xi_k\}_{k\in \N_0}$ are independent $N(0_d,I_d)$ and $\nabla\cdot(b\rho)=\partial_t \rho$, and we defined 
\begin{equation}
\label{eq:update}
\begin{aligned}    \zeta_{k}(x,y)&=U_k(x)-U_{k+1}(y)+\frac{1}{2}(x-y)\cdot \left(\frac{[b_k(x)-b_{k+1}(y)]}{\eps}+[s_k(x)+s_{k+1}(y)]\right)\\&+\frac{h}{4}\left(\frac{|b_k(x)|^2-|b_{k+1}(y)|^2}{\eps}+\eps [|s_k(x)|^2-|s_{k+1}(y)|^2]\right)+\\&+\frac{h}{2}(b_k(x)\cdot s_k(x)+b_{k+1}(y)\cdot s_{k+1}(y))
\end{aligned}
\end{equation}
Then, for all $k\in \N_0$,
\begin{equation}
\label{eq:grad:Z:k:bis}
    \E_{\theta_k} [\partial_{\theta} U_{\theta_k}]=\frac{\E[  \partial_{\theta} U_{\theta_k}(X_k)e^{A_k} ]}{\E [ e^{A_k}]}, \qquad Z_{\theta_k} = Z_{\theta_0} \E \left[e^{A_k}\right]
\end{equation}
where the expectations on the right-hand side are over the law of the joint process $(X_k,A_k)$.
\end{prop}

\begin{proof}
The proof is a generalization of the one of Proposition \ref{prop:3}. The iteration rule for $A_k$ in \eqref{eq:X:A:k:bis} implies that
\begin{equation}
\label{eq:X:A:k:s:bis}
A_k \coloneqq \sum_{q=1}^k \left(\alpha_{q-1}(X_{q-1},X_q) - \lambda_{q}(X_{q},X_{q-1})\right), \qquad k\in \N.
\end{equation}
where

\begin{equation}
\label{eq:alpha:bis}
    \alpha_{k}(x,y)=U(t_k,x)-\frac{1}{2\eps}(y-x)\cdot [b_k(x)+\eps s_k(x)]+\frac{h}{4\eps}|b_k(x)+\eps s_k(x)|^2
\end{equation}
and
\begin{equation}
\label{eq:lambda}
    \lambda_{k}(x,y)=U(t_k,x)-\frac{1}{2\eps}(y-x)\cdot [-b_k(x)+\eps s_k(x)]+\frac{h}{4\eps}|-b_k(x)+\eps s_k(x)|^2
\end{equation}
For $k\in \N_0$, let 
\begin{equation}
    \pi^{forw}_k(x,y)=(2\pi h\eps)^{-d/2}\exp\Big(-\frac{1}{4h\eps}\left|y-x-h[b_k(x)+\eps s_k(x)]\right|^2\Big)
\end{equation}
be the transition probability density of the ULA update in~\eqref{eq:X:A:k:bis}, i.e. $\pi^{forw}_k(X_{k},X_{k+1})$ is the probability density of $X_{k+1}$ conditionally on $X_k$. Similarly,
\begin{equation}
    \pi^{back}_k(x,y)=(2\pi h\eps)^{-d/2}\exp\Big(-\frac{1}{4h\eps}\left|y-x-h[-b_k(x)+\eps s_k(x)]\right|^2\Big)
\end{equation}
Notice the minus sign before $b$ in the backward transition kernel. By the definition of $A_k$, we have
\begin{equation}
\label{eq:X:A:k:s2:bis}
\begin{aligned}
    \exp(A_k) &= \prod_{q=1}^k \exp\left(\alpha_{q-1}(X_{q-1},X_q) - \lambda_{q}(X_{q},X_{q-1})\right)\\
    & =  e^{-U_{\theta_k}(X_k)+U_{\theta_0}(X_0)} \prod_{q=1}^k \frac{ \pi^{back}_q(X_q,X_{q-1})}{\pi^{forw}_{q-1}(X_{q-1},X_{q})}
\end{aligned}
\end{equation}
where in the second line we added and subtracted $|X_q - X_{q-1}|^2/4h\eps$ and used the definition of $\alpha_k(x,y)$ and $\lambda_k(x,y)$ given in~\eqref{eq:alpha:bis} and ~\eqref{eq:lambda}. 
Since the joint probability density function of the path $(X_0,X_1,\ldots,X_k)$ at any $k\in \N$ is
\begin{equation}
\label{eq:varrho:bis}
    \varrho(x_0,x_1,\ldots, x_k) = Z_{\theta_0}^{-1} e^{-U_{\theta_0}(x_0)} \prod_{q=1}^k \pi^{forw}_{q-1}(x_{q-1},x_{q})
\end{equation}
we deduce from~\eqref{eq:X:A:k:s2:bis} and \eqref{eq:varrho:bis} that, given an $f:\R^d \to \R$,  we can express the expectation $\E[f(X_k) e^{A_k}]$ as the integral
\begin{equation}
    \label{eq:expect:int:bis}
    \begin{aligned}
    &\E[f(X_k) e^{A_k}]\\
    &= \int_{\R^{dk}} f(x_k) e^{-U_{\theta_k}(x_k)+U_{\theta_0}(x_0)} \prod_{q=1}^k \frac{ \pi^{back}_q(x_q,x_{q-1})}{\pi^{forw}_{q-1}(x_{q-1},x_{q})} \varrho(x_0,x_1,\ldots, x_k) dx_0\cdots dx_k\\
    &  = Z_{\theta_0}^{-1} \int_{\R^{dk}} f(x_k) e^{-U_{\theta_k}(x_k)} \prod_{q=1}^k  \pi^{back}_q(x_q,x_{q-1})dx_0\cdots dx_k
    \end{aligned}
\end{equation}
Since $\int_{\R^d} \pi_k(x,y) dy =1$ for all $k\in\N_0$ and all $x\in\R^d$, we can perform the integrals over $x_0$, then $x_1$, etc. in this expression to be left with
\begin{equation}
    \label{eq:expect:int:2:bis}
    \E[f(X_k) e^{A_k}]= Z_{\theta_0}^{-1} \int_{\R^{d}} f(x_k) e^{-U_{\theta_k}(x_k)} dx_k
\end{equation}
Setting $f(x)=1$ in this expression gives
\begin{equation}
    \label{eq:f1:bis}
    \E[e^{A_k}]= Z_{\theta_0}^{-1} \int_{\R^{d}} e^{-U_{\theta_k}(x_k)} dx_k = Z_{\theta_0}^{-1} Z_{\theta_k}
\end{equation}
which implies Jarzynski identity; combining \eqref{eq:expect:int:2:bis} and 
\eqref{eq:f1:bis} we obtain the fundamental relation
\begin{equation}
    \E_{t_k} [f]=\frac{\E[  f(X_k)e^{A_k} ]}{\E [ e^{A_k}]}
\end{equation}
\end{proof}
\begin{Lemma}
\label{lem:triv}
    If the SDE is integrated with Euler-Maruyama method of $O(h^{3/2})$, then 
    \begin{equation}
        A_{k+1}-A_k=O(h^{3/2})
    \end{equation}
    that is, for $h\to 0$ we have $A_k\equiv 0$ for any $k\geq 0$.
\end{Lemma}
\begin{proof}
    Expanding $A_{k+1}-A_k$ in series of $h$ we notice that $O(h^{1/2})$ vanishes since
\begin{equation}
\begin{aligned}
    A_{k+1}-A_k&=\underbrace{-\partial_t U_k(X_k) h\textcolor{black}{+(X_{k+1}-X_k)\cdot s_k(X_k)}\textcolor{black}{+\frac{1}{2}(X_{k+1}-X_k)^T J^s_k(X_k) (X_{k+1}-X_k) }}_{U_k(X_k)-U_{k+1}(X_{k+1})}  
    \\&\textcolor{black}{-(X_{k+1}-X_k)\cdot  s_k(X_k)}+\frac{1}{2\eps}(X_{k+1}-X_k)^T J^b_k(X_k)(X_{k+1}-X_k)\\
    &\textcolor{black}{-\frac{1}{2}(X_{k+1}-X_k)^T J^s_k(X_k)(X_{k+1}-X_k)} +h b_k(X_k)\cdot s_k(X_k) +O(h^{3/2})
\end{aligned}
\end{equation}
where $J_b$ and $J_s$ are the Jacobian of $b$ and $s$. After simplifications we get 
\begin{equation}
\begin{aligned}
    A_{k+1}-A_k&=-\partial_t U_k(X_k) h+\frac{1}{2\eps}(X_{k+1}-X_k)^T J^b_k(X_k)(X_{k+1}-X_k)+\\&+h b_k(X_k)\cdot s_k(X_k)+O(h^{3/2})
    \end{aligned}
\end{equation}
We can use \eqref{3} and $\xi_k^T J^b_k(X_k)\xi_k \stackrel{d}{=} \nabla \cdot b_k(X_k)$ (in law) to further manipulate the expression
    \begin{equation}
    \begin{aligned}
    A_{k+1}-A_k=-\partial_t U_k(X_k) h+h b_k(X_k)\cdot s_k(X_k)+h\nabla \cdot b_k(X_k)+ O(h^{3/2})    
    \end{aligned}
\end{equation}
In conclusion, we notice how  $\partial_t\rho_t(x)+\nabla\cdot (b(t,x)\rho_t(x))=0$ implies
\begin{equation}
    -\partial_t U_k(X_k) + b_k(X_k)\cdot s_k(X_k)+\nabla \cdot b_k(X_k)=0
\end{equation}
that yields to the expected $A_{k+1}-A_k=O(h^{3/2})$. 
\end{proof}
The presented results show that we have no Jarzynski correction in the continuous limit. On the contrary, for any finite time step $h$, the weights have a non trivial evolution, meaning that their presence is necessary to correct time discretization error. Hence, we proved how Jarzynski reweighting can be beneficial in a more general context, that is the generation phase for flow-based methods \eqref{3}. Plus, we expect Lemma \ref{lem:triv} to generalize: if one adopts an higher-order integrator for the SDE then the expansion for $h\to 0$ of $A_{k+1}-A_k$ will be of the same order, e.g. with error $O(h^{l/2})$ with $l>3$ on the integrator we will obtain $A_{k+1}-A_k=O(h^{l/2})$.
\subsection{Jarzynski Reweighting for Restricted Boltzmann Machines}
\label{sec:rbm}
Restricted Boltzmann Machines (RBMs) are a class of Energy-Based Models that operate over a discrete state space and are characterized by a bipartite architecture. Originally introduced as a simplified variant of Boltzmann Machines \cite{ackley1985learning}, RBMs gained prominence due to their efficient training via contrastive divergence \cite{hinton2002training}. Their probabilistic structure, rooted in statistical mechanics, has made them a fundamental tool in generative modeling, feature learning, and dimensionality reduction \cite{salakhutdinov2009deep}. In this section we will present the structure of the transition kernel for two important classes of RBM, in order to clarify their possible use in the context of Proposition \ref{prop:3}.\\
The first class we consider are the Bernoulli RBM. Such a model consists of a set of  $d_v$  visible units,  $v = (v_1, \dots, v_{d_v})$ , representing observed data, and a set of  $d_h$  hidden units,  $h = (h_1, \dots, h_{d_h})$ , which capture latent features. Both visible and hidden variables are binary, taking values in  $\{0,1\}$ , leading to a discrete configuration space of size  $2^{d_v + d_h}$. The energy function defining the probability distribution of the system is given by
\begin{equation}
U_\theta(v, h) = - \sum_{\alpha=1}^{d_v} b_\alpha v_\alpha - \sum_{\beta=1}^{d_h} c_\beta h_\beta - \sum_{\alpha=1}^{d_v} \sum_{\beta=1}^{d_h} v_\alpha W_{\alpha\beta} h_\beta.
\end{equation}
This expression highlights the structural constraints of the model. The first two terms describe independent biases applied to each visible and hidden unit, respectively, while the last term captures the interaction between visible and hidden layers through the weight matrix  W . The bipartite nature of the model is explicit: connections exist only between the two layers, with no intra-layer interactions. The discreteness of the state space is crucial in shaping the statistical properties of the RBM. Unlike models with continuous variables, where the energy landscape can exhibit smooth deformations, the RBM operates over a combinatorial space where probability mass is concentrated on a discrete set of points. This feature significantly impacts both the sampling dynamics and the learning process, as updates must navigate a rugged energy landscape constrained by the binary nature of the variables.\\
Since the training is a cross-entropy minimization task, the problem of sampling from $\rho_\theta$ is present also in this case. The main difference is that the state space is discrete -- one cannot use Langevin-like sampling since it is defined on a continuous state space. A possible method is the so-called Gibbs sampling \cite{fischer2012introduction}.
\begin{defi}[Gibbs Sampling for Bernoulli RBMs]
    We use \(\sigma(x)\) as the sigmoid function:
$\sigma(x) = 1/(1 + e^{-x}$. 
Given the current state $\{v,h\}$ of an RBM, the new state $\{v',h'\}$ is obtained in the following way:
    \begin{enumerate}
    \item 
The conditional probability of hidden unit \( h_i = 1 \) for $i=1,\dots,d_h$ given the visible units \( v \) is 
   \begin{equation}
       P(h'_i = 1 \mid v) = \sigma\left(c_i + \sum_{j=1}^m v_j W_{ji}\right)
   \end{equation}
   Then, sample using a Bernoulli distribution $h'_i\sim \text{Bernoulli}(P(h'_i = 1 \mid v))$

\item The conditional probability of visible unit \( v_j = 1 \) for $j=1,\dots,d_v$ given the visible units \( h' \) is
   \begin{equation}
       P(v_j = 1 \mid h') = \sigma\left(b_j + \sum_{i=1}^n h'_i W_{ji}\right)
   \end{equation}
   Then, sample using a Bernoulli distribution $v'_j\sim \text{Bernoulli}(P(v'_j = 1 \mid h'))$
\end{enumerate}
By alternating between these two steps—sampling \( h \) given \( v \), and then sampling \( v \) given \( h \)—the system converges to the joint distribution \( \rho_\theta(v, h) \), cfr. \cite{zhang2018overview}. See also Appendix \ref{app:rbm} for more details.
\end{defi}
Similarly to MCMC routines for standard EBMs, Gibbs sampling is used to obtain sample from $\rho_\theta$; but it is an {\it equilibrium} sampler -- $(v,h)$ converges to a sample from $\rho_\theta$ only after infinitely many iterations. Thus, computing the gradient of cross-entropy is affected by the same bias discussed in Section \ref{sec:ebm}. Hence, we present a generalization of Proposition \ref{prop:3} in case of Bernoulli RBM.
\begin{Lemma}
\label{lem:rbm}
    Consider a Bernoulli RBM; using the notation of Proposition \ref{prop:3} Gibbs sampling induces a transition kernel $\pi(X_{k+1},X_k)\equiv \pi((v', h'),(v,h))$ such that:
\begin{equation}
\label{43}
    \begin{aligned}
\log \pi((v', h'),(v,h)) &= \sum_{i=1}^n \left[ h'_i \log\left(\sigma\left( c_i + \sum_{j=1}^m v_j W_{ji} \right)\right) + \right.\\
&\left.+(1 - h'_i) \log\left(1 - \sigma\left( c_i + \sum_{j=1}^m v_j W_{ji} \right)\right) \right]+
\\
&+ \sum_{j=1}^m \left[ v'_j \log\left(\sigma\left( b_j + \sum_{i=1}^n h'_i W_{ji} \right)\right) +\right.\\
&\left.+ (1 - v'_j) \log\left(1 - \sigma\left( b_j + \sum_{i=1}^n h'_i W_{ji} \right)\right) \right]
\end{aligned}
\end{equation}
\end{Lemma}
\begin{proof}
    The log-probabilities for the transition kernel describe the log-probability of moving from one state to another. The log-probability of \( h' \) given \( v \) is:
\begin{equation}
    \begin{aligned}
        \log P(h' \mid v) &= \sum_{i=1}^n \left[ h'_i \log\left(\sigma\left( c_i + \sum_{j=1}^m v_j W_{ji} \right)\right) + \right.\\
&\left. +(1 - h'_i) \log\left(1 - \sigma\left( c_i + \sum_{j=1}^m v_j W_{ji} \right)\right) \right]
    \end{aligned}
\end{equation}

Similarly, the log-probability of \( v' \) given \( h' \) is:
\begin{equation}
    \begin{aligned}
\log P(v' \mid h') &= \sum_{j=1}^m \left[ v'_j \log\left(\sigma\left( b_j + \sum_{i=1}^n h'_i W_{ji} \right)\right) + \right.\\
&\left.+(1 - v'_j) \log\left(1 - \sigma\left( b_j + \sum_{i=1}^n h'_i W_{ji} \right)\right) \right]
    \end{aligned}
\end{equation}
In both expression we just used Bernoulli probability distribution, layer-wise independence and basic properties of logarithm. Since the two sampling steps are subsequential, the transition kernel is the product of the conditional probabilities -- that is, we can sum the two log-probabilities obtaining the log-transition kernel \eqref{43}.
\end{proof}
The second model we consider is a Gaussian-Bernoulli Restricted Boltzmann Machine; it represents an hybrid situation between standard EBMs and discrete RBM since the visible units are continuous-valued, while the hidden units remain binary \cite{cho2011improved}. The energy function is given by:

\begin{equation}
U_\theta(v, h) = \sum_{\alpha=1}^{d_v} \frac{(v_\alpha - b_\alpha)^2}{2\sigma_\alpha^2} - \sum_{\beta=1}^{d_h} c_\beta h_\beta - \sum_{\alpha=1}^{d_v} \sum_{\beta=1}^{d_h} \frac{v_\alpha}{\sigma_\alpha} W_{\alpha\beta} h_\beta.
\end{equation}
In this case, the parameter set $\theta = \{W, b, c, \sigma^2\}$ includes not only the weight matrix and biases but also the variance parameters $\sigma^2\in\mathbb{R}^{d_v}_+$, which account for different noise levels in the visible units. This modification allows the model to handle continuous data more effectively by incorporating Gaussian noise in the visible layer.

\begin{defi}[Gibbs Sampling for Gaussian RBMs]
Given the current state $\{v,h\}$ of an RBM, the new state $\{v',h'\}$ is obtained in the following way:
\begin{enumerate}
\item The conditional probability of hidden unit \( h_i = 1 \) for $i=1,\dots,d_h$ given the visible units \( v \) is
\begin{equation}
P(h'_i = 1 \mid v) = \sigma\left(c_i + \sum_{j=1}^{d_v} \frac{v_j}{\sigma_j} W_{ji} \right),
\end{equation}
where $\sigma$ is the sigmoid function. Then, sample using a Bernoulli distribution $h'_i\sim \text{Bernoulli}(P(h'_i = 1 \mid v))$.

\item The conditional distribution of visible unit \( v_j \) for \( j=1,\dots,d_v \) given the hidden units \( h' \) follows a Gaussian $\mathcal{N}$:
\begin{equation}
    P(v'_j \mid h') = \mathcal{N} \left( b_j + \sum_{i=1}^{d_h} W_{ji} h'_i,\, \sigma_j^2 \right).
\end{equation}
Then, sample \( v'_j \sim \mathcal{N}(b_j + \sum_{i=1}^{d_h} W_{ji} h'_i, \sigma_j^2) \).

\end{enumerate}
By alternating these two steps, the system converges to the joint distribution $\rho_\theta$.
\end{defi}
\begin{Lemma}
Consider a Gaussian-Bernoulli RBM. The Gibbs sampling process induces a transition kernel  such that:
\begin{equation}
\begin{aligned}
\log \pi((v', h'),(v,h)) &= \sum_{i=1}^{d_h} \left[ h'_i \log\left(\sigma\left( c_i + \sum_{j=1}^{d_v} \frac{v_j}{\sigma_j} W_{ji} \right)\right) +\right. \
\\& \left. + (1 - h'_i) \log\left(1 - \sigma\left( c_i + \sum_{j=1}^{d_v} \frac{v_j}{\sigma_j} W_{ji} \right)\right) \right] \
+\\& - \sum_{j=1}^{d_v} \frac{(v'_j - b_j - \sum_{i=1}^{d_h} W_{ji} h'_i)^2}{2\sigma_j^2} - \sum_{j=1}^{d_v} \log \left( \sqrt{2\pi} \sigma_j \right)
\end{aligned}
\end{equation}
\end{Lemma}

\begin{proof}
The proof is similar to Bernoulli RBM: the transition kernel consists of two independent steps. The log-probability of hiddent neurons given visible follows the Bernoulli conditional probability:
\begin{equation}
\begin{aligned}
\log P(h' \mid v) &= \sum_{i=1}^{d_h} \left[ h'_i \log\left(\sigma\left( c_i + \sum_{j=1}^{d_v} \frac{v_j}{\sigma_j} W_{ji} \right)\right)\right.\\&\left. + (1 - h'_i) \log\left(1 - \sigma\left( c_i + \sum_{j=1}^{d_v} \frac{v_j}{\sigma_j} W_{ji} \right)\right) \right].
\end{aligned}
\end{equation}
Similarly, since each \( v'_j \) is sampled from a normal distribution, its log-probability density function is:
\begin{equation}
\log P(v' \mid h') = - \sum_{j=1}^{d_v} \frac{(v'_j - b_j - \sum_{i=1}^{d_h} W_{ji} h'_i)^2}{2\sigma_j^2} - \sum_{j=1}^{d_v} \log \left( \sqrt{2\pi} \sigma_j \right).
\end{equation}
Summing the log-probabilities from these two steps yields the final result.
\end{proof}
\section{Discussion and Future Perpectives}
In conclusion, we have demonstrated that Jarzynski reweighting remains applicable even when different transition kernels — i.e., different sampling dynamics for $\rho_\theta$ — are used. This flexibility is crucial for effectively computing the gradient of the cross-entropy and represents an important generalization compared to the original formulation.\\The significance of this generalization lies in the fact that, as discussed in \cite{carbone2024efficient,carbone2024generative} and in the broader context of non-equilibrium sampling, reweighting is a delicate process. Typically, resampling is required to control the effective sample size, which can be highly sensitive to the choice of dynamics governing the state variables. To illustrate this in intuitive terms, consider a finite population of samples $\{X^i\}_{i=1}^N$. The reweighting process adjusts the importance of individual samples when computing expectations such as \eqref{eq:grad:Z:k} through empirical weighted averaging. The extent to which a sample contributes to the estimate depends on whether it follows the evolving distribution $\rho_\theta$. If the empirical variance of the weights is large, only a few samples will significantly contribute to the estimate, reducing statistical efficiency and increasing the risk of high variance in the gradient estimation.\\
To mitigate this issue, it is essential to design sampling dynamics that are well-suited for training. Ideally, in a perfectly chosen dynamics, the importance weights would remain constant and uniform across samples \cite{vaikuntanathan2008escorted} since the latter are perfectly following the evolving density; and this avoids the need for frequent resampling. Thus, understanding how different transition kernels influence the evolution of weights is a key aspect of applying Jarzynski reweighting effectively.\\
In this report, we examined two distinct scenarios that are representative of major approaches in generative modeling. First, we considered diffusion-based flow models, which are among the most advanced generative techniques. In Subsection \ref{sec:ula}, we showed that during the generative phase—i.e., while integrating the associated stochastic differential equation (SDE)—Jarzynski reweighting can be employed to correct numerical errors introduced by the discretization scheme. Second, in Subsection \ref{sec:rbm}, we analyzed the use of Jarzynski reweighting in the training of Restricted Boltzmann Machines (RBMs), a fundamental class of Energy-Based Models (EBMs). RBMs are not only relevant from a practical standpoint but also hold historical significance in the development of generative models.\\
These results emphasize the versatility of Jarzynski reweighting in different generative modeling frameworks and highlight the importance of selecting an appropriate transition kernel to ensure robust and efficient training dynamics. Future work could further investigate optimal kernel selection strategies and explore broader applications of reweighting techniques in modern generative models, in order to scale up the method to applications.

\section*{Acknowledgements}
 D.C. has worked under the auspices of Italian National Group of Mathematical
Physics (GNFM). D.C. was supported by the Italian Ministry of University and Research (MUR) through the grant “Ecosistemi dell’innovazione”, costruzione di “leader territoriali di R\&S” with grant agreement no. ECS00000036. D.C. thanks Eric Vanden-Eijnden and Lamberto Rondoni for the fruitful discussions and for the support.
 \bibliographystyle{unsrtnat}
 \bibliography{reference}

 \newpage
 \appendix

\input{main_appendix_camera_ready.tex}


\end{document}

%% file: main_appendix_camera_ready.tex
\section{Jarzynski reweighting in continuous-time}
\label{app:continuous}

We give a continuous-time version of Jarzynski reweighting, taken from \cite{carbone2024efficient}. 
\begin{prop}
\label{prop:2}
Assume that the parameters $\theta$ are evolved according to some time-differentiable protocol $\theta(t)$ such that $\theta(0)=\theta_0$. Given any $\alpha>0$, let $X_t\in \R^d$ and $A_t\in\R$ be the solutions of
\begin{equation}
\label{eq:X:A:t}
\left\{ 
\begin{aligned}
    dX_t&=-\alpha \nabla U_{\theta(t)}(X_t) dt+\sqrt{2\alpha}\,dW_t,\quad\quad &&X_0\sim \rho_{\theta_0},\\
    \dot A_t&=-\partial_\theta{U}_{\theta(t)}(X_t)\cdot \dot \theta(t),   &&A_0=0.
\end{aligned}\right.  
\end{equation}
where $U_{\theta}(x)$ is the model energy and $W_t\in\R^d$ is a standard Wiener process.
Then, for any $t\ge0$,
\begin{equation}
\label{eq:grad:Z:t}
    \E_{\theta(t)} [\partial_\theta U_{\theta(t)}]=\frac{\E[  \partial_\theta U_{\theta(t)}(X_t)e^{A_t}]}{\E [ e^{A_t}]}, \qquad Z_{\theta(t)} = Z_{\theta_0} \E [ e^{A_t}],
\end{equation}
where the expectations on the right-hand side are over the law of the joint process $(X_t,A_t)$.
\end{prop}
The second equation in~\eqref{eq:grad:Z:t} can also be written in term of the free energy $F_\theta = - \log Z_\theta$ as $F_{\theta(t)}  = F_{\theta_0}-\log \E [e^{A_t}]$: this is  Jarzynski's equality~\cite{jarzynski1997nonequilibrium}. We stress that it is key to include the weights in~\eqref{eq:grad:Z:t} and, in particular, $\E[  \partial_\theta U_{\theta(t)}(X_t)] \not=\E_{\theta(t)} [\partial_\theta U_{\theta(t)}]$. This is because the PDF of $X_t$ alone lags behind the model PDF $\rho_{\theta(t)}$: the larger $\alpha$, the smaller this lag, but it is always there if $\alpha<\infty$,  see the remark at the end of this section for more discussion on this point. The inclusion of the weights in~\eqref{eq:grad:Z:t} corrects exactly for the bias induced by this lag.

An immediate consequence of Proposition~\ref{prop:2} is that we can evolve $\theta(t)$ by the gradient descent flow over the cross-entropy by solving~\eqref{eq:X:A:t} concurrently with
\begin{equation}
\label{eq:GD:t}
    \dot \theta(t)=\dfrac{\E[\partial_\theta U_{\theta(t)}(X_t)e^{A_t}]}{\E[e^{A_t}]}-\E_*[\partial_\theta U_{\theta(t)}],\quad\quad\quad \quad\theta(0)=\theta_0
\end{equation}
since by~\eqref{eq:grad:Z:t} the right hand side of~\eqref{eq:GD:t} is precisely $-\partial_\theta H(\rho_{\theta(t)}, \rho_*) = \E_{\theta(t)} [\partial_\theta U_{\theta(t)}] - \E_*[\partial_\theta U_{\theta(t)}]$. Assuming that we know $Z_{\theta_0}$ we can also track the evolution of the cross-entropy \eqref{2} via
\begin{equation}
    \label{eq:c:e:time}
    H(\rho_{\theta(t)}, \rho_*) = \log \E [ e^{A_t}] + \log Z_{\theta_0} + \E_* [U_{\theta(t)}].
\end{equation}

\begin{proof}
The joint PDF $f(t,x,a)$ of the process $(X_t,A_t)$ satisfying~\eqref{eq:X:A:t} solves the Fokker-Planck equation (FPE)
\begin{equation}
\label{11b}
     \partial_t f=\alpha\nabla_x\cdot(\nabla_x U_{\theta(t)} f +\nabla_x f)+\partial_\theta{U}_{\theta(t)}\cdot\dot\theta(t) \partial_af, \quad f(0,x,a) = Z_{\theta_0}^{-1} e^{-U_{\theta_0}(x)} \delta(a).
\end{equation} 
Let us derive an equation for 
\begin{equation}
    \hat \rho(t,x) = \int_{-\infty}^\infty e^a f(t,x,a) da
\end{equation}
To this end, multiply~\eqref{11b} by~$e^a$, integrate the result over $a\in(-\infty,\infty)$, and use integration by parts for the last term at the right-hand side to obtain:
\begin{equation}
\label{eq:hatrho}
     \partial_t \hat \rho =\alpha \nabla_x\cdot(\nabla_x U_{\theta(t)} \hat \rho +\nabla_x \hat \rho)-\partial_\theta{U}_{\theta(t)}\cdot\dot\theta(t) \hat \rho, 
     \qquad \hat \rho(0,x) = Z_{\theta_0}^{-1} e^{-U_{\theta_0}(x)}
\end{equation}
By general results for the solutions of parabolic PDEs such as \eqref{eq:hatrho} (see \cite{evans2022partial}, Chapter 7), we know that the solution to this equation is unique, and 
we can check by direct substitution that it is given by 
\begin{equation}
\label{eq:hatrho:sol}
     \hat \rho(t,x) = Z_{\theta_0}^{-1} e^{-U_{\theta(t)}(x)}.
\end{equation}
This implies that 
\begin{equation}
\label{eq:hatrho:sol:int}
     \int_{\R^d} \hat \rho(t,x) dx   = Z_{\theta_0}^{-1} Z_{\theta(t)}.
\end{equation}
Since by definition $\E[e^{A_t}] = \int_{\mathbb{R}^d}\int_{-\infty}^\infty e^a f(t,x,a) da dx = \int_{\R^d} \hat \rho(t,x) dx$ this establishes the second equation in~\eqref{eq:grad:Z:t} . To establish the first notice that
\begin{equation}
    \label{eq:exp:f}
    \begin{aligned}
    \frac{\E[ \partial_\theta U_{\theta(t)}(X_t)e^{A_t}]}{\E [e^{A_t}]} &= \frac{\int_{\mathbb{R}^d}\int_{-\infty}^\infty   \partial_\theta U_{\theta(t)}(x) e^a  f(t,x,a)  dadx}{\int_{\mathbb{R}^d}\int_{-\infty}^\infty e^a f(t,x,a) dadx}= \frac{\int_{\R^d} \partial_\theta U_{\theta(t)}(x) \hat \rho(t,x)dx}{\int_{\R^d} \hat \rho(t,x) dx} \\
    &= \frac{Z_{\theta_0}^{-1}\int_{\mathbb{R}^d}\partial_\theta U_{\theta(t)}(x)e^{-U_{\theta(t)}(x)}dx}{Z_{\theta_0}^{-1}Z_{\theta(t)} } = \E_{\theta(t)} [\partial_\theta U_{\theta(t)}]
    \end{aligned}
\end{equation} 

\end{proof}

\paragraph{The need for Jarzynski's correction.}
Suppose that the walkers satisfy the Langevin equation (first equation in~\eqref{eq:X:A:t}):
\begin{equation}
    \label{eq:langevin}
     dX_t=-\alpha \nabla U_{\theta(t)}(X_t) dt+\sqrt{2\alpha}\,dW_t,\quad\quad X_0\sim \rho_{\theta_0}, 
\end{equation}
where $\theta(t)$ is evolving according to some protocol.
The probability density function $\rho(t,x)$ of $X_t$ then satisfies the Fokker-Planck equation (compare~\eqref{eq:hatrho})
\begin{equation}
    \label{eq:fpe:l}
    \partial_t \rho = \alpha\nabla\cdot \left(\nabla U_{\theta(t)}(x) \rho + \nabla \rho\right), \qquad \rho(t=0) = \rho_{\theta_0}
\end{equation}
The solution to this equation is not available in closed form, and in particular $\rho(t,x) \not = \rho_{\theta(t)}(x)$ --  $\rho(t,x)$ is only close to $\rho_{\theta(t)}(x)$ if we let $\alpha\to\infty$, so that the walkers $X_t$ move much faster than the parameters $\theta(t)$ but this limit is not easily achievable in practice (as convergence of the FPE solution to its equilibrium is very slow in general if the potential $U_{\theta(t)}$ is complicated). As a result $\E[ \partial_\theta U_{\theta(t)}(X_t)] \not =\E_{\theta(t)} [\partial_\theta U_{\theta(t)}]$, implying the necessity to include the weights in the expectation~\eqref{eq:grad:Z:t}. As a side note, \eqref{eq:langevin} is formally equivalent to PCD in continuous time, up to a possible different choice for the initial condition $X_0$ -- sometimes it is sampled from another chosen density relying on the fact that \eqref{eq:langevin} should converge anyway to a law coincident with $\rho_\theta$. 

\section{Contrastive divergence and persistent contrastive divergence algorithms}
\label{app:cd:pcd}
Both Constrastive Divergence and Persistent Constrastive Divergence try to generate the necessary samples for estimating $\mathbb{E}_\theta$ using a Markov Chain. Without loss of generality, we focus on the choice of Unadjusted Langevin Algorithm, cfr. Algorithms 3 and 4 in \cite{carbone2024efficient}. Considering continuous time, the samples are subject to the Langevin equation:
\begin{equation}
     dX_t=-\alpha \nabla U_{\theta(t)}(X_t) dt+\sqrt{2\alpha}\,dW_t,\quad\quad X_0\sim \rho_{\theta_0}, 
\end{equation}
where $\theta(t)$ is evolving according to some learning protocol.
The probability density function $\rho(t,x)$ of $X_t$ then satisfies the Fokker-Planck equation (compare~\eqref{eq:hatrho})
\begin{equation}
    \partial_t \rho = \alpha\nabla\cdot \left(\nabla U_{\theta(t)}(x) \rho + \nabla \rho\right), \qquad \rho(t=0) = \rho_{\theta_0}
\end{equation}
The solution to this equation is not available in closed form, and in particular $\rho(t,x) \not = \rho_{\theta(t)}(x)$ --  $\rho(t,x)$ is only close to $\rho_{\theta(t)}(x)$ if we let $\alpha\to\infty$, so that the walkers $X_t$ move much faster than the parameters $\theta(t)$ but this limit is not easily achievable in practice (as convergence of the FPE solution to its equilibrium is very slow in general if the potential $U_{\theta(t)}$ is complicated). As a result $\E[ \partial_\theta U_{\theta(t)}(X_t)] \not =\E_{\theta(t)} [\partial_\theta U_{\theta(t)}]$, implying the necessity to include the weights in the expectation~\eqref{eq:grad:Z:t}.

Interestingly, we can write down an equation that mimics the evolution of the PDF of the walkers in the CD algorithm, at least in the continuous-time limit: this equation reads
\begin{equation}
    \label{eq:fpe:pcd}
    \partial_t \check\rho = \alpha\nabla\cdot \left(\nabla U_{\theta(t)}(x) \check\rho + \nabla \check\rho\right) - \nu (\check\rho-\rho_*), \qquad \check\rho(t=0) = \rho_{*}
\end{equation}
where the parameter $\nu>0$ controls the rate at which the walkers are reinitialized at the data points: the last term in~\eqref{eq:fpe:pcd} is a birth-death term that captures the effect of these reinitializations. The solution to this equation is not available in closed from (and $\check\rho(t,x) \not = \rho_{\theta(t)}(x)$ in general), but in the limit of large~$\nu$ (i.e. with very frequent reinitializations), we can show~\cite{domingoenrich2022dual} that
\begin{equation}
    \label{eq:large:nu}
    \check\rho(t,x) = \rho_*(x) + \nu^{-1} \alpha\nabla\cdot \left(\nabla U_{\theta(t)}(x) \rho_*(x) + \nabla \rho_*(x)\right) + O(\nu^{-2}).
\end{equation}
As a result
\begin{equation}
    \label{eq:cd:expect} 
    \begin{aligned}
        &\int_{\R^d} \partial_\theta U_{\theta(t)}(x) (\rho_*(x)-\check \rho(t,x)) dx\\
        &= -\nu^{-1} \int_{\R^d} \partial_\theta U_{\theta(t)}(x) \nabla\cdot \left(U_{\theta(t)}(x) \rho_*(x) + \nabla \rho_*(x)\right)dx + O(\nu^{-2})\\
        & = \nu^{-1} \int_{\R^d} \left( \partial_\theta \nabla U_{\theta(t)}(x) \cdot \nabla U_{\theta(t)}(x) - \partial_\theta \Delta U_{\theta(t)}(x)\right) \rho_*(x) dx + O(\nu^{-2})
    \end{aligned}
\end{equation}
The leading order term at the right hand side is precisely $\nu^{-1}$ times the gradient with respect to $\theta$ of the Fisher divergence
\begin{equation}
    \label{eq:fisher}
    \begin{aligned}
        &\frac12\int_{\R^d} |\nabla U_{\theta}(x) + \nabla \log\rho_*(x)|^2\rho_*(x) dx \\
        & = \frac12\int_{\R^d} \left[ |\nabla U_{\theta}(x)|^2 -2 \Delta U_{\theta}(x) +| \nabla \log\rho_*(x)|^2\right]\rho_*(x) dx
        \end{aligned}
\end{equation}
where $\Delta$ denotes the Laplacian and we used $\int_{\R^d} \nabla U_{\theta}(x) \cdot \nabla \log\rho_*(x) \rho_*(x) dx= \int_{\R^d} \nabla U_{\theta}(x) \cdot \nabla \rho_*(x) dx = - \int_{\R^d} \Delta U_{\theta}(x)  \rho_*(x) dx$. This confirms the known fact that the CD algorithm effectively performs GD on the Fisher divergence rather than the cross-entropy \cite{hyvarinen2007connections}. 

\section{Derivation of the Conditional Probability in an RBM}
\label{app:rbm}

In this appendix, we derive the conditional probability of a hidden unit given the visible units in a Restricted Boltzmann Machine (RBM). Specifically, we show that

\begin{equation}
    P(h_i = 1 \mid v) = \sigma\left(c_i + \sum_{j=1}^{m} v_j W_{ji} \right),
\end{equation}

where \(\sigma(x) = \frac{1}{1 + e^{-x}}\) is the sigmoid function. Due to the bipartite structure of the RBM, the hidden units are conditionally independent given the visible units. The conditional probability can be written as:

\begin{equation}
    P(h_i = 1 \mid v) = \frac{P(h_i = 1, v)}{P(v)}.
\end{equation}

From the Boltzmann distribution, we have:

\begin{equation}
    P(h_i = 1 \mid v) = \frac{e^{-U(v, h_i=1)}}{e^{-U(v, h_i=0)} + e^{-U(v, h_i=1)}}.
\end{equation}

Using the energy function, the difference in energy between the two cases is:

\begin{equation}
    U(v, h_i=0) - U(v, h_i=1) = c_i + \sum_{j=1}^{m} v_j W_{ji}.
\end{equation}

Thus, the conditional probability simplifies to:

\begin{equation}
    P(h_i = 1 \mid v) = \frac{e^{-(U(v, h_i=1) - U(v, h_i=0))}}{1 + e^{-(U(v, h_i=1) - U(v, h_i=0))}}.
\end{equation}

Since the sigmoid function is defined as:

\begin{equation}
    \sigma(x) = \frac{1}{1 + e^{-x}},
\end{equation}

we obtain:

\begin{equation}
    P(h_i = 1 \mid v) = \sigma\left(c_i + \sum_{j=1}^{m} v_j W_{ji} \right).
\end{equation}

This result follows from the fact that the energy function is linear in the binary variables \( h_i \), and when conditioning on the visible units, the probability distribution over hidden units follows a logistic (sigmoid) form. In a similar fashion one can derive an analogue result for $P(v_i = 1 \mid h)$.

%% file: neurips_camera_ready.bbl
\begin{thebibliography}{51}
\providecommand{\natexlab}[1]{#1}
\providecommand{\url}[1]{\texttt{#1}}
\expandafter\ifx\csname urlstyle\endcsname\relax
  \providecommand{\doi}[1]{doi: #1}\else
  \providecommand{\doi}{doi: \begingroup \urlstyle{rm}\Url}\fi

\bibitem[LeCun et~al.(2007)LeCun, Chopra, and Hadsell]{lecun2006tutorial}
Yann LeCun, Sumit Chopra, and Raia Hadsell.
\newblock A tutorial on energy-based learning.
\newblock In G{\"o}khan BakIr, Thomas Hofmann, Alexander~J Smola, Bernhard Sch{\"o}lkopf, and Ben Taskar, editors, \emph{Predicting structured data}, chapter~10. MIT press, 2007.

\bibitem[Gutmann and Hyv{\"a}rinen(2010)]{gutmann2010noise}
Michael Gutmann and Aapo Hyv{\"a}rinen.
\newblock Noise-contrastive estimation: A new estimation principle for unnormalized statistical models.
\newblock In \emph{Proceedings of the thirteenth international conference on artificial intelligence and statistics}, pages 297--304. JMLR Workshop and Conference Proceedings, 2010.

\bibitem[Song et~al.(2020)Song, Garg, Shi, and Ermon]{song2020sliced}
Yang Song, Sahaj Garg, Jiaxin Shi, and Stefano Ermon.
\newblock Sliced score matching: A scalable approach to density and score estimation.
\newblock In \emph{Uncertainty in Artificial Intelligence}, pages 574--584. PMLR, 2020.

\bibitem[Mehrabi et~al.(2021)Mehrabi, Morstatter, Saxena, Lerman, and Galstyan]{mehrabi2021survey}
Ninareh Mehrabi, Fred Morstatter, Nripsuta Saxena, Kristina Lerman, and Aram Galstyan.
\newblock A survey on bias and fairness in machine learning.
\newblock \emph{ACM Computing Surveys (CSUR)}, 54\penalty0 (6):\penalty0 1--35, 2021.

\bibitem[Hyv{\"a}rinen and Dayan(2005)]{hyvarinen2005estimation}
Aapo Hyv{\"a}rinen and Peter Dayan.
\newblock Estimation of non-normalized statistical models by score matching.
\newblock \emph{Journal of Machine Learning Research}, 6\penalty0 (4), 2005.

\bibitem[Hinton(2002)]{hinton2002training}
Geoffrey~E Hinton.
\newblock Training products of experts by minimizing contrastive divergence.
\newblock \emph{Neural computation}, 14\penalty0 (8):\penalty0 1771--1800, 2002.

\bibitem[Welling and Hinton(2002)]{welling2002new}
Max Welling and Geoffrey~E Hinton.
\newblock A new learning algorithm for mean field boltzmann machines.
\newblock In \emph{International conference on artificial neural networks}, pages 351--357, 2002.

\bibitem[Carreira-Perpinan and Hinton(2005)]{carreira2005contrastive}
Miguel~A Carreira-Perpinan and Geoffrey Hinton.
\newblock On contrastive divergence learning.
\newblock In \emph{International workshop on artificial intelligence and statistics}, pages 33--40. PMLR, 2005.

\bibitem[Hyvarinen(2007)]{hyvarinen2007connections}
Aapo Hyvarinen.
\newblock Connections between score matching, contrastive divergence, and pseudolikelihood for continuous-valued variables.
\newblock \emph{IEEE Transactions on neural networks}, 18\penalty0 (5):\penalty0 1529--1531, 2007.

\bibitem[Carbone et~al.(2024{\natexlab{a}})Carbone, Hua, Coste, and Vanden-Eijnden]{carbone2024efficient}
Davide Carbone, Mengjian Hua, Simon Coste, and Eric Vanden-Eijnden.
\newblock Efficient training of energy-based models using jarzynski equality.
\newblock \emph{Advances in Neural Information Processing Systems}, 36, 2024{\natexlab{a}}.

\bibitem[Carbone et~al.(2024{\natexlab{b}})Carbone, Hua, Coste, and Vanden-Eijnden]{carbone2024generative}
Davide Carbone, Mengjian Hua, Simon Coste, and Eric Vanden-Eijnden.
\newblock Generative models as out-of-equilibrium particle systems: Training of energy-based models using non-equilibrium thermodynamics.
\newblock In \emph{International Conference on Nonlinear Dynamics and Applications}, pages 287--311. Springer, 2024{\natexlab{b}}.

\bibitem[Song and Kingma(2021)]{song2021train}
Yang Song and Diederik~P Kingma.
\newblock How to train your energy-based models.
\newblock \emph{arXiv preprint arXiv:2101.03288}, 2021.

\bibitem[Xie et~al.(2016)Xie, Lu, Zhu, and Wu]{xie2016theory}
Jianwen Xie, Yang Lu, Song-Chun Zhu, and Yingnian Wu.
\newblock A theory of generative convnet.
\newblock In \emph{International Conference on Machine Learning}, pages 2635--2644. PMLR, 2016.

\bibitem[Grathwohl et~al.(2019)Grathwohl, Wang, Jacobsen, Duvenaud, Norouzi, and Swersky]{grathwohl2019your}
Will Grathwohl, Kuan-Chieh Wang, Joern-Henrik Jacobsen, David Duvenaud, Mohammad Norouzi, and Kevin Swersky.
\newblock Your classifier is secretly an energy based model and you should treat it like one.
\newblock In \emph{International Conference on Learning Representations}, 2019.

\bibitem[Brooks et~al.(2011)Brooks, Gelman, Jones, and Meng]{brooks2011handbook}
Steve Brooks, Andrew Gelman, Galin Jones, and Xiao-Li Meng.
\newblock \emph{Handbook of Markov chain Monte Carlo}.
\newblock CRC press, 2011.

\bibitem[Liu and Liu(2001)]{liu2001monte}
Jun~S Liu and Jun~S Liu.
\newblock \emph{Monte Carlo strategies in scientific computing}, volume~75.
\newblock Springer, 2001.

\bibitem[Vincent(2011)]{vincent2011connection}
Pascal Vincent.
\newblock A connection between score matching and denoising autoencoders.
\newblock \emph{Neural computation}, 23\penalty0 (7):\penalty0 1661--1674, 2011.

\bibitem[Swersky et~al.(2011)Swersky, Ranzato, Buchman, Freitas, and Marlin]{swersky2011autoencoders}
Kevin Swersky, Marc'Aurelio Ranzato, David Buchman, Nando~D Freitas, and Benjamin~M Marlin.
\newblock On autoencoders and score matching for energy based models.
\newblock In \emph{International conference on machine learning (ICML-11)}, pages 1201--1208, 2011.

\bibitem[Wenliang(2022)]{wenliang2022failure}
Li~Kevin Wenliang.
\newblock On the failure of variational score matching for {VAE} models.
\newblock \emph{arXiv preprint arXiv:2210.13390}, 2022.

\bibitem[Wenliang et~al.(2019)Wenliang, Sutherland, Strathmann, and Gretton]{wenliang2019learning}
Li~Wenliang, Danica~J Sutherland, Heiko Strathmann, and Arthur Gretton.
\newblock Learning deep kernels for exponential family densities.
\newblock In \emph{International Conference on Machine Learning}, 2019.

\bibitem[Song and Ermon(2019)]{song2019generative}
Yang Song and Stefano Ermon.
\newblock Generative modeling by estimating gradients of the data distribution.
\newblock In \emph{Advances in neural information processing systems}, volume~32, 2019.

\bibitem[Tieleman(2008)]{tieleman2008training}
Tijmen Tieleman.
\newblock Training restricted boltzmann machines using approximations to the likelihood gradient.
\newblock In \emph{International conference on Machine learning}, pages 1064--1071, 2008.

\bibitem[Xie et~al.(2018)Xie, Lu, Gao, Zhu, and Wu]{xie2018cooperative}
Jianwen Xie, Yang Lu, Ruiqi Gao, Song-Chun Zhu, and Ying~Nian Wu.
\newblock Cooperative training of descriptor and generator networks.
\newblock \emph{IEEE transactions on pattern analysis and machine intelligence}, 42\penalty0 (1):\penalty0 27--45, 2018.

\bibitem[Nijkamp et~al.(2019)Nijkamp, Hill, Zhu, and Wu]{nijkamp2019learning}
Erik Nijkamp, Mitch Hill, Song-Chun Zhu, and Ying~Nian Wu.
\newblock Learning non-convergent non-persistent short-run {MCMC} toward energy-based model.
\newblock In \emph{Advances in Neural Information Processing Systems}, volume~32, 2019.

\bibitem[Gao et~al.(2020)Gao, Nijkamp, Kingma, Xu, Dai, and Wu]{gao2020flow}
Ruiqi Gao, Erik Nijkamp, Diederik~P Kingma, Zhen Xu, Andrew~M Dai, and Ying~Nian Wu.
\newblock Flow contrastive estimation of energy-based models.
\newblock In \emph{Proceedings of the IEEE/CVF Conference on Computer Vision and Pattern Recognition}, pages 7518--7528, 2020.

\bibitem[Jarzynski(1997)]{jarzynski1997nonequilibrium}
C~Jarzynski.
\newblock Nonequilibrium equality for free energy differences.
\newblock \emph{Physical Review Letters}, 78\penalty0 (14):\penalty0 2690, 1997.

\bibitem[Neal(2001)]{neal2001annealed}
Radford~M Neal.
\newblock Annealed importance sampling.
\newblock \emph{Statistics and computing}, 11:\penalty0 125--139, 2001.

\bibitem[Doucet et~al.(2001)Doucet, De~Freitas, Gordon, et~al.]{doucet2001sequential}
Arnaud Doucet, Nando De~Freitas, Neil~James Gordon, et~al.
\newblock \emph{Sequential Monte Carlo methods in practice}, volume~1.
\newblock Springer, 2001.

\bibitem[Le et~al.(2018)Le, Igl, Rainforth, Jin, and Wood]{anh2018autoencoding}
Tuan~Anh Le, Maximilian Igl, Tom Rainforth, Tom Jin, and Frank Wood.
\newblock Auto-encoding sequential monte carlo.
\newblock In \emph{International Conference on Learning Representations}, 2018.
\newblock URL \url{https://openreview.net/forum?id=BJ8c3f-0b}.

\bibitem[Ding and Freedman(2020)]{ding2020learning}
Xinqiang Ding and David~J. Freedman.
\newblock Learning deep generative models with annealed importance sampling, 2020.

\bibitem[Midgley et~al.(2023)Midgley, Stimper, Simm, Sch{\"o}lkopf, and Hern{\'a}ndez-Lobato]{midgley2023flow}
Laurence~Illing Midgley, Vincent Stimper, Gregor N.~C. Simm, Bernhard Sch{\"o}lkopf, and Jos{\'e}~Miguel Hern{\'a}ndez-Lobato.
\newblock Flow annealed importance sampling bootstrap.
\newblock In \emph{The Eleventh International Conference on Learning Representations}, 2023.
\newblock URL \url{https://openreview.net/forum?id=XCTVFJwS9LJ}.

\bibitem[Du et~al.(2023)Du, Durkan, Strudel, Tenenbaum, Dieleman, Fergus, Sohl-Dickstein, Doucet, and Grathwohl]{du2023reduce}
Yilun Du, Conor Durkan, Robin Strudel, Joshua~B Tenenbaum, Sander Dieleman, Rob Fergus, Jascha Sohl-Dickstein, Arnaud Doucet, and Will~Sussman Grathwohl.
\newblock Reduce, reuse, recycle: Compositional generation with energy-based diffusion models and mcmc.
\newblock In \emph{International Conference on Machine Learning}, pages 8489--8510. PMLR, 2023.

\bibitem[Sohl-Dickstein et~al.(2015)Sohl-Dickstein, Weiss, Maheswaranathan, and Ganguli]{pmlr-v37-sohl-dickstein15}
Jascha Sohl-Dickstein, Eric Weiss, Niru Maheswaranathan, and Surya Ganguli.
\newblock Deep unsupervised learning using nonequilibrium thermodynamics.
\newblock In Francis Bach and David Blei, editors, \emph{Proceedings of the 32nd International Conference on Machine Learning}, volume~37 of \emph{Proceedings of Machine Learning Research}, pages 2256--2265, Lille, France, 07--09 Jul 2015. PMLR.
\newblock URL \url{https://proceedings.mlr.press/v37/sohl-dickstein15.html}.

\bibitem[Doucet et~al.(2022)Doucet, Grathwohl, Matthews, and Strathmann]{doucet2022scorebased}
Arnaud Doucet, Will~Sussman Grathwohl, Alexander G. D.~G. Matthews, and Heiko Strathmann.
\newblock Score-based diffusion meets annealed importance sampling.
\newblock In \emph{Advances in Neural Information Processing Systems}, 2022.

\bibitem[Salakhutdinov and Hinton(2009)]{salakhutdinov2009deep}
Ruslan Salakhutdinov and Geoffrey Hinton.
\newblock Deep boltzmann machines.
\newblock In \emph{Artificial intelligence and statistics}, pages 448--455. PMLR, 2009.

\bibitem[Hinton(2012)]{hinton2012practical}
Geoffrey~E Hinton.
\newblock A practical guide to training restricted boltzmann machines.
\newblock \emph{Neural Networks: Tricks of the Trade: Second Edition}, pages 599--619, 2012.

\bibitem[Cho et~al.(2011)Cho, Ilin, and Raiko]{cho2011improved}
KyungHyun Cho, Alexander Ilin, and Tapani Raiko.
\newblock Improved learning of gaussian-bernoulli restricted boltzmann machines.
\newblock In \emph{Artificial Neural Networks and Machine Learning--ICANN 2011: 21st International Conference on Artificial Neural Networks, Espoo, Finland, June 14-17, 2011, Proceedings, Part I 21}, pages 10--17. Springer, 2011.

\bibitem[Lipman et~al.(2022)Lipman, Chen, Ben-Hamu, Nickel, and Le]{lipmanflow}
Yaron Lipman, Ricky~TQ Chen, Heli Ben-Hamu, Maximilian Nickel, and Matthew Le.
\newblock Flow matching for generative modeling.
\newblock In \emph{The Eleventh International Conference on Learning Representations}, 2022.

\bibitem[Albergo et~al.(2023)Albergo, Boffi, and Vanden-Eijnden]{albergo2023stochastic}
Michael~S Albergo, Nicholas~M Boffi, and Eric Vanden-Eijnden.
\newblock Stochastic interpolants: A unifying framework for flows and diffusions.
\newblock \emph{arXiv preprint arXiv:2303.08797}, 2023.

\bibitem[Oksendal(2003)]{oksendal2003stochastic}
Bernt Oksendal.
\newblock \emph{Stochastic Differential Equations}.
\newblock Springer-Verlag Berlin Heidelberg, 6 edition, 2003.

\bibitem[Mattingly et~al.(2002)Mattingly, Stuart, and Higham]{mattingly2002stochastic}
J.~C. Mattingly, A.~M. Stuart, and D.~J. Higham.
\newblock {Ergodicity for SDEs and approximations: locally Lipschitz vector fields and degenerate noise}.
\newblock \emph{Stochastic Processes and their Applications}, 101\penalty0 (2):\penalty0 185--232, October 2002.

\bibitem[Talay and Tubaro(1990)]{talay1900expansion}
Denis Talay and Luciano Tubaro.
\newblock Expansion of the global error for numerical schemes solving stochastic differential equations.
\newblock \emph{Stochastic Analysis and Applications}, 8\penalty0 (4):\penalty0 483--509, 1990.

\bibitem[Carbone(2024)]{carbphd}
Davide Carbone.
\newblock Generative models as out-of-equilibrium particle systems: the case of energy-based models.
\newblock 2024.

\bibitem[Parisi(1981)]{parisi1981correlation}
Giorgio Parisi.
\newblock Correlation functions and computer simulations.
\newblock \emph{Nuclear Physics B}, 180\penalty0 (3):\penalty0 378--384, 1981.

\bibitem[Durmus and Moulines(2017)]{durmus2017nonasymptotic}
Alain Durmus and {\'E}ric Moulines.
\newblock Nonasymptotic convergence analysis for the unadjusted langevin algorithm.
\newblock \emph{THE ANNALS of APPLIED PROBABILITY}, pages 1551--1587, 2017.

\bibitem[Vaikuntanathan and Jarzynski(2008)]{vaikuntanathan2008escorted}
Suriyanarayanan Vaikuntanathan and Christopher Jarzynski.
\newblock Escorted free energy simulations: Improving convergence by reducing dissipation.
\newblock \emph{Physical Review Letters}, 100\penalty0 (19):\penalty0 190601, 2008.

\bibitem[Ackley et~al.(1985)Ackley, Hinton, and Sejnowski]{ackley1985learning}
David~H Ackley, Geoffrey~E Hinton, and Terrence~J Sejnowski.
\newblock A learning algorithm for boltzmann machines.
\newblock \emph{Cognitive science}, 9\penalty0 (1):\penalty0 147--169, 1985.

\bibitem[Fischer and Igel(2012)]{fischer2012introduction}
Asja Fischer and Christian Igel.
\newblock An introduction to restricted boltzmann machines.
\newblock In \emph{Progress in Pattern Recognition, Image Analysis, Computer Vision, and Applications: 17th Iberoamerican Congress, CIARP 2012, Buenos Aires, Argentina, September 3-6, 2012. Proceedings 17}, pages 14--36. Springer, 2012.

\bibitem[Zhang et~al.(2018)Zhang, Ding, Zhang, and Xue]{zhang2018overview}
Nan Zhang, Shifei Ding, Jian Zhang, and Yu~Xue.
\newblock An overview on restricted boltzmann machines.
\newblock \emph{Neurocomputing}, 275:\penalty0 1186--1199, 2018.

\bibitem[Evans(2022)]{evans2022partial}
Lawrence~C Evans.
\newblock \emph{Partial differential equations}, volume~19.
\newblock American Mathematical Society, 2022.

\bibitem[Domingo-Enrich et~al.(2021)Domingo-Enrich, Bietti, Gabri{\'e}, Bruna, and Vanden-Eijnden]{domingoenrich2022dual}
Carles Domingo-Enrich, Alberto Bietti, Marylou Gabri{\'e}, Joan Bruna, and Eric Vanden-Eijnden.
\newblock Dual training of energy-based models with overparametrized shallow neural networks.
\newblock \emph{arXiv preprint arXiv:2107.05134}, 2021.

\end{thebibliography}
